\documentclass[review]{elsarticle}

\usepackage{lineno,hyperref}

\usepackage{todonotes}
\usepackage{amsmath,amssymb}

\journal{Journal of Biomedical Informatics}

\begin{document}

\begin{frontmatter}
\title{Single versus Multiple Annotation for Named Entity Recognition of Mutations}


\author{David Martinez Iraola, Antonio Jimeno Yepes}
\address{IBM Research, Southbank, VIC, Australia}

\begin{abstract}

The focus of this paper is to address the knowledge acquisition bottleneck for Named Entity Recognition (NER) of mutations, by analysing different approaches to build manually-annotated data. We address first the impact of using a single annotator vs two annotators, in order to measure whether multiple annotators are required. Once we evaluate the performance
loss when using a single annotator, we apply different methods to sample
the training data for second annotation, aiming at improving the quality of the dataset without requiring a full pass. We use held-out double-annotated data to build two scenarios with different types of rankings:
similarity-based and confidence based. We evaluate both approaches on: (i) their ability to
identify training instances that are erroneous (cases where single-annotator labels
differ from double-annotation after discussion), and (ii) on Mutation NER performance for state-of-the-art classifiers after integrating the fixes at different thresholds.


\end{abstract}

\begin{keyword}
\texttt{Natural Language Processing}\sep \texttt{Information Extraction}\sep
\texttt{Named Entity Recognition}\sep
\texttt{Data Acquisition}
\end{keyword}

\end{frontmatter}


\section{Introduction}

Natural Language Processing (NLP) is a research field that is
receiving increasing attention, because of its potential to transform
the bulk of written human knowledge into actionable structured
data. One example of this potential is applied to research in
the area of DNA mutations, whose findings are mainly distributed
through publications describing experiments and their
outcomes. Reading the large number of new articles has become an
unmanageable task, and NLP tools are being used to automatically
extract information from the research literature. This information can
be represented in databases or knowledge bases, and allow researchers
to more efficiently access knowledge.

Mutations are curated from the scientific literature into databases
such as COSMIC~\cite{forbes2010cosmic}, which are valuable resources in clinical
pipelines. Since these resources are curated from the literature,
automating the extraction of mutation information using text
analytics will speed up the transfer of knowledge into structured databases.
Genetic mutations are usually represented in text using standard nomenclature,
such as the one defined by the HGVS (Human Genome Variation Society)
\footnote{\url{https://varnomen.hgvs.org}}. Many tools
have used regular expressions, or a combination of regular expressions
and machine learning to identify these mentions in the literature~\cite{nagel2009annotation,caporaso2007mutationfinder,jimeno2014mutation}.
In addition to these structured representations, mutations are as well
mentioned in the literature using free natural language~\cite{jimeno2018hybrid}.

One limitation of using NLP tools is the lack of generic solutions
that can be applied to text in multiple domains for extracting
information. In particular for mutation extraction, existing
Biomedical NLP tools (MetaMap, etc.) will not cover certain kinds of
phenomena, and state-of-the-art pre-trained models such as BERT~\cite{devlin2018bert} require training data from the
domain, which is hard to obtain.
In the mutation domain,  \citet{jimeno2018hybrid} developed
annotation and tools to extract different types of mentions and
related entities. They relied on a group of domain-expert annotators,
and found that high performance can be achieved by using deep
learning. Their work also showed that an important bottleneck to
achieve high performance was the cost of manual annotation, and they
illustrated the importance of multiple annotators per instance to
increase annotation quality. Manual annotation inconsistencies or errors can have large effects in
performance, and they are difficult to identify and fix once the annotation
process is finished. The best approach when resources are available (used in~\cite{jimeno2018hybrid}) consists on multiple annotators going over the same texts separately, and discussing when there are disagreements. However, often this approach is not feasible, and a single annotator is used for most of the annotation, with the intervention of a second one over a sample, to provide inter-annotator agreement statistics. Some widely used datasets for domain-specific NLP follow this approach, for instance BIONLP-2013-GENIA~\cite{kim-etal-2013-genia}, or NICTA-PIBOSO~\cite{kim2011automatic}.

The goal of this paper is to address the knowledge acquisition bottleneck by analysing different approaches to build manually-annotated data. We focus first on the cost of using a single annotator vs two annotators, in order to measure whether there is a large gap in results between the two options. Once we evaluate the performance
loss using a single annotator, we apply different methods to sample
the training data for second annotation, aiming at improving the quality of the dataset. We use held-out double-annotated data to build two scenarios with different types of rankings:
similarity-based and confidence based. We evaluate both methods on: (i) their ability to
identify training instances that are erroneous (cases where single-annotator labels
differ from double-annotation), and (ii) on Mutation NER performance for state-of-the-art classifiers after integrating the fixes at different thresholds. This work is related to active learning~\cite{settles2009active}, which aims at effective sampling of unlabeled data for annotation. The difference is that in our case we focus on comparing single vs double annotation in a
challenging domain, and on providing methods to improve over single annotation with targeted second annotation.

The rest of the article is organised as follows. We present related work in Section~\ref{sec:related work}, and then introduce the experimental setting in Section~\ref{sec:experimental setting}. The results are explained in Section~\ref{sec:results}, and we provide discussion and conclusions in Sections~\ref{sec:discussion} and \ref{sec:conclusion}.



\section{Related work}
\label{sec:related work}

Manual annotation is a time consuming and expensive activity, but required to train supervised machine learning methods, and to tune or evaluate algorithms.
Previous work on annotating mutation related entities~\cite{verspoor2013annotating,jimeno2018hybrid} has relied on manual annotation from several annotators that is used to measure inter-annotator agreement, but as well to improve the quality of single annotated corpora.

There are several methods considered to improve the speed or cost at which manual annotation is collected.
One of such methods is active learning~\cite{settles2009active}, the idea being to achieve high accuracy with fewer training examples, by posing queries in the form of unlabeled instances to be labeled by an oracle (typically a human annotator). The instances are chosen using techniques to optimise the performance of the classifiers. In our case the number of annotated instances is fixed, and we apply methods to identify a sample to re-annotate in order to improve the performance of the models.





The cost is not the only factor that affects manual annotation outcomes.
Disagreements between human annotators might affect the quality of manually annotated sets.
This is clearly a problem in Mechanical Turk (MT) settings~\cite{rodrigues2013learning}, in which random annotators need to be identified and discarded. There are also approaches such as \textit{task routing} where instance difficulty is modeled, and appropriate annotators are chosen for each instance~\cite{yang-etal-2019-predicting}, leading to improved datasets. When multiple annotators per instance are available, methods to integrate annotations from multiple sources such as in~\textit{repeated labelling}~\cite{ipeirotis2014repeated, rodrigues2014gaussian} (e.g. using MT) provide higher performance. In our work, annotators have been trained during the generation of the guidelines and annotation inconsistencies are due to the complexity of the domain (they may be as well linked to domain experience), and errors might be caused by overlooking entities present in the documents.


There is related work in the area of automatically fixing annotation errors. In \cite{abaho2019correcting} manual rules and machine learning are combined to automatically re-annotate an existing public dataset, leading to improved models built on the data.

\section{Experimental setting}
\label{sec:experimental setting}

The goal of the experimental pipeline is to explore the following research questions:
\begin{enumerate}
\item What is the performance difference between using single-annotated versus double-annotated data to build models?
\item Can we reduce the gap between single and double-annotated data-based models by using similarity and confidence-based methods to select instances for double annotation?
\end{enumerate}

We rely on the IBM-Mutation dataset (Section~\ref{sec:dataset}) to build and evaluate two
classifiers based on deep learning: BioBERT~\cite{lee2020biobert}, and Bilstm-crf~\cite{tran2017named}. Both
classifiers are state of the art, and they provide log probabilities
with their predictions, which we use these as confidence scores.

\begin{figure}[htbp]
\centerline{\includegraphics[scale=.5]{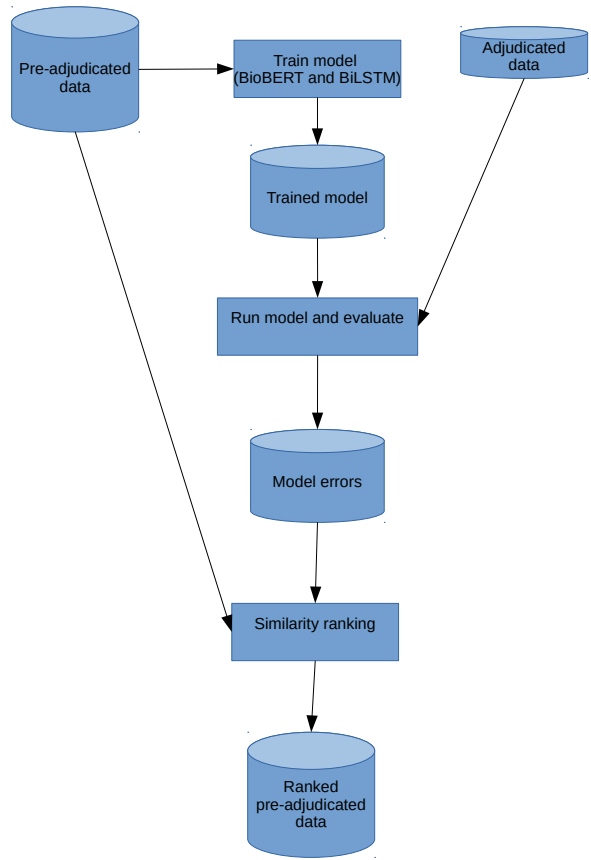}}
\caption{Similarity-based scenario. Training data is ranked according to their similarity to error models over development data.}
\label{fig:similarity_flowchart}
\end{figure}

We define two scenarios to evaluate methods that can improve the
manual annotation process. In the first scenario (Figure~\ref{fig:similarity_flowchart}), we use both a
small adjudicated dataset, and a larger pre-adjudicated dataset. The
pre-adjudicated dataset is employed to train the initial classifiers
that are tested over the small adjudicated dataset, and the instances
with classification errors point to similar instances in the training
data via textual similarity methods. The intuition in this case is
that the single-annotated training examples that are similar to test
sentences producing errors, are more likely to be erroneous
themselves.

\begin{figure}[htbp]
\centerline{\includegraphics[scale=.5]{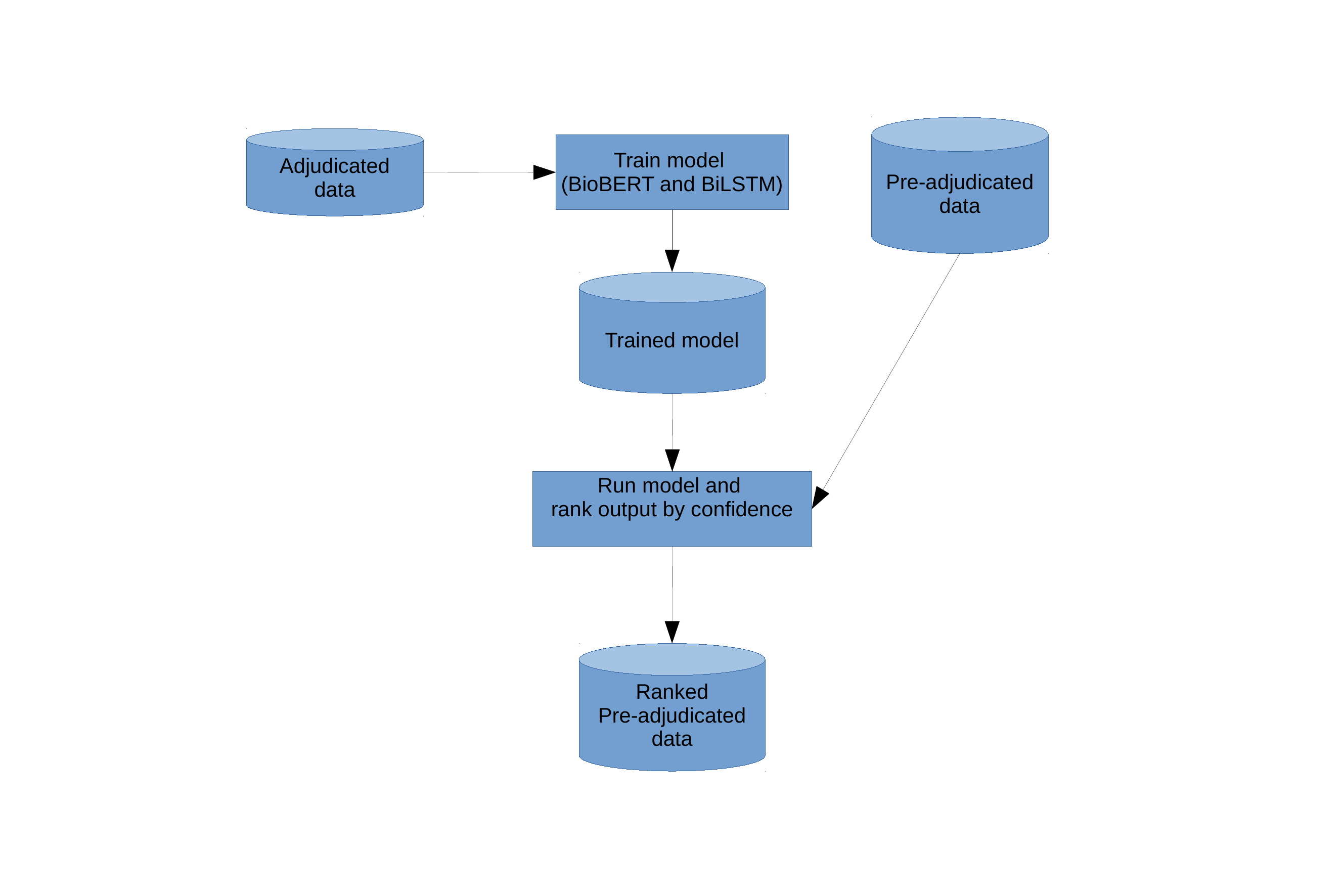}}
\caption{Confidence-based scenario. Training data is ranked according to the confidence of the classifiers.}
\label{fig:confidence_flowchart}
\end{figure}

In the second scenario (Figure~\ref{fig:confidence_flowchart}) we rely on a small sample of the data that is
double-annotated, and we build the initial classifiers by training the models on
this dataset. Then we run the classifiers over the rest of the data,
and rank all sentences according to the confidence scores. The
intuition is that the sentences with lower confidence are the most
likely to be difficult to annotate, and therefore prone to errors.

We describe the classifiers we have used in Section~\ref{sec:classifiers}, the sentence similarity approaches required for the first scenario in Section~\ref{sentence similarity}, and the dataset in Section~\ref{sec:dataset}.

\subsection{Classifiers}
\label{sec:classifiers}

\subsubsection{Bilstm-crf}

We have used a Bilstm-crf method~\cite{tran2017named} that is comprised of an encoding mechanism that uses a bidirectional LSTM (Long Short Term Memory) network~\cite{graves2013speech}, which acts as a feature encoding that substitutes manual engineering of features, and a decoding method that uses a neural CRF~\cite{lample2016neural} to generate the named entity recognition annotation tags.


        
Three layers of bidirectional LSTM have been stacked using residual connections. The LSTM dimension we used is 100, running 100 epochs with a dropout rate of 0.5. All the other parameters are the same as in previous work~\cite{tran2017named}. Named entity recognition is a sequence labeling task. In Bilstm-crf, the probability of a sequence is shown in equation~\ref{eq:soft}. The score $s(Y,S)$, as explained in~\cite{tran2017named} is used in softmax with the scores of all possible sequences $\sum_{Y'\in \mathbb{Y}} e^{s(Y',S)}$.

We use the output of equation~\ref{eq:soft} to obtain the confidence of the Bilstm-crf classifier.


\begin{align}
	\label{eq:soft}
	Pr(Y|S) = \frac{e^{s(Y,S)}}{\sum_{Y'\in \mathbb{Y}} e^{s(Y',S)}}
\end{align}

\subsubsection{BioBERT}

BioBERT is a domain specific pre-training of BERT (Bidirectional Encoder Representations from Transformers)~\cite{devlin2018bert}, which uses English Wikipedia, BooksCorpus, PubMed abstracts and PubMed Central (PMC) full text articles as corpus. We relied on pre-trained weights from the combination of BioBERT Base 1.0 with added PubMed and PMC documents.
For NER fine-tuning, BioBERT has a single output layer that uses the output of the last BERT layer to predict BIO2 NER labels. Due to the random initialization of BERT, different results can be expected for each run; in order to reduce the variability we average results over 10 runs, and provide the standard deviation when using BioBERT.

The average of the logits from the output layer are used as the confidence of the prediction of the BioBERT NER system.


\subsection{Sentence Similarity}
\label{sentence similarity}


Semantic textual similarity (STS) is a widely studied problem, and multiple approaches exist~\cite{cer-etal-2017-semeval}. One of the most successful systems in recent public evaluations is based on word alignments across the input texts~\cite{sultan2015dls}. This method is unsupervised, and it uses the Paraphrase Database (PPDB)~\cite{ganitkevitch-etal-2013-ppdb} to identify semantically similar words, by computing dependencies and surface-form neighbors of the two words to determine their contextual similarity. We rely on the implementation by ~\citet{brychcin-svoboda-2016-uwb} to apply this approach. Apart from the similarity score, this method provides alignments between text snippets in the different sentences, which can be used to provide an interpretation of the score. This is an example of interpretable semantic textual similarity (ISTS), and we use this term to refer to this method.

As a different approach, similarity methods based on word embeddings have proven successful over shared datasets and tasks~\cite{reimers2019sentence}. In the biomedical domain, recent work has shown the robust performance of BioBERT~\cite{molla2020query}, and our second similarity method will rely on this technique, by averaging the word embeddings obtained before running the last layer of BioBERT.


\subsection{Dataset}
\label{sec:dataset}

In this study, we have used the IBM-Mutation corpus presented in~\cite{jimeno2018hybrid} and publicly available\footnote{\url{https://github.com/ibm-aur-nlp/amia-18-mutation-corpus}}.
The annotation was performed by 5 domain experts on MEDLINE citations related to colorectal cancer. During the manual annotation process two annotators worked on the same document independently and disagreements were afterwards resolved. 
The data set contains 167 MEDLINE citations and it contains 60 mentions of \textit{DNA modifications}, 1324 mentions of \textit{genes/proteins}, 320 mentions of \textit{loci}, 1337 mentions of \textit{mutations} and 23 mentions of \textit{RNAs}.
The data set contains 94 \textit{component of} relations, 52 \textit{has modification} relations and 907 \textit{has mutation} relations. For our experiments we focus on entities of type \textit{mutation}.

The IBM-Mutation corpus
has been double-annotated, and each sentence has been separately annotated by two experts. We rely on this annotation to build two versions of the dataset: pre-adjudicated (single annotation), and adjudicated (agreed labels after discussion).
For our experiments, when an instance has different labels in the different
versions, we consider the pre-adjudicated label to be erroneous. For instance, in the adjudicated dataset the following sentence contains the annotations in bold, both of which are missing in the pre-adjudicated dataset: "Sixteen of 17 \textbf{mutations} were at residue 599 (\textbf{V599E})." We incorporated the partial annotations to the existing distribution of the IBM-Mutation dataset for the research community.


For our experiments, we randomly split the data into 4 groups: training, development, test-held-out (test1), and test
(test2). The number of sentences and annotations are given in Table~\ref{tab:dataset}. We initially split the dataset into training and test (80\%/20\%); then further split each part randomly into two similar sets, to be used as trainining$/$development and test-held-out$/$test. The table does not provide pre-annotated cases for test, since they are not used (we evaluate on the adjudicated labels only). We can see that the adjudicated corpus contains more annotations, meaning that the single annotation tends to miss cases.


\begin{table}[ht]
  \centering
  \caption{Dataset splits used for the experiments.}
  \label{tab:dataset}
  \begin{tabular}{|llll|}
    \hline
    \textbf{Split}    & \textbf{Sentences} & \textbf{Annotations} & \textbf{Annotations}\\ 
    &&\textbf{Pre-adjudicated}&\textbf{Adjudicated}\\
    \hline
Training & 717 & 823 & 927\\
Development & 614 & 489 & 616\\
Test-held-out (test1) & 167 & - & 124\\
Test (test2) & 195 & - & 226\\
    \hline
  \end{tabular}
\end{table}

\section{Results}
\label{sec:results}

We first compute the performance of different NER settings in Section~\ref{sec:NER_performance}, and then we evaluate the scores of ranking methods to detect erroneous sentences in Section~\ref{sec:ranking eval}. Finally in Section~\ref{sec:retraining} we re-train and test the models with different thresholds of double-annotated instances.

\subsection{NER classifier performance}
\label{sec:NER_performance}

Our first goal is to measure the performance of NER classifiers over
the IBM-Mutation dataset, and compare the results to the state of the
art. The results at the bottom of Table~\ref{ner_results} show the performance of
our two classifiers over test data (test2) when trained over
double-annotated (adjudicated) training data. For comparison with the state of the art,
we include some relevant performances reported over this dataset
by~\cite{jimeno2018hybrid}. Direct comparison is not possible, due to different data splits, but the results illustrate that our BioBERT classifier is competitive, and even scores close to the reported inter-annotator agreement for the dataset.

\begin{table}[ht]
  \centering
  \caption{Performance of NER systems for Mutation NER.}
  \label{ner_results}
  \begin{tabular}{|llll|}
    \hline
    \textbf{System}    & \textbf{Precision} & \textbf{Recall}&\textbf{F-score (Std Dev)}\\ \hline
Bilstm-crf(\cite{jimeno2018hybrid})&82.0&68.8&74.9\\
Inter-annotator agreement(\cite{jimeno2018hybrid})&79.0&77.6&78.3\\
Hybrid system(\cite{jimeno2018hybrid})&85.0&74.8&79.6\\
\hline
Bilstm-crf&72.0&72.4&72.2\\
BioBERT & 76.1 & 76.1 & 76.0 (2.1)\\
\hline
  \end{tabular}
\end{table}


We next measure the scores when only single-annotated
data is used to train the models. We test our classifiers using the same test set (test2), and the
results are given in Table~\ref{ner_results_preadj}. We can see that in this scenario the performance drops, specially for BioBERT (-9.8\%), and less catastrophically for Bilstm-crf (-4.6\%). This is the first indication of the importance of second annotators when building NER datasets.

\begin{table}[t]
  \centering
  \caption{Performance of NER systems for Mutation NER using single-labeled data for training.}
  \label{ner_results_preadj}
  \begin{tabular}{|llll|}
    \hline
    \textbf{System}    & \textbf{Precision} & \textbf{Recall}&\textbf{F-score (Std Dev)}\\ \hline
BioBERT& 73.4 & 63.3 & 67.9 (2.8)\\
Bilstm-crf&71.4&64.1&67.6\\
\hline
  \end{tabular}
\end{table}


\subsection{Finding annotation errors in single-labeled data}
\label{sec:ranking eval}

In order to reduce the gap in performance between training with double
and single-labeled data, we explore the possibility of selectively double-annotating training
instances to improve the model, without having to double-annotate the
full dataset. There are different methods to identify the best
instances for re-annotation, and in this work we focus on confidence
and similarity-based techniques. We first explore the ability of these
methods to identify single-annotated cases in training data that change labels when they go over second annotation.

Our first step is to evaluate similarity-based methods (cf. Figure~\ref{fig:similarity_flowchart}). For this experiment, we run models that are trained with pre-adjudicated data over the held-out test data
(test1), and identify the errors committed by the model. These errors are used as guide to find instances to fix in training data, via similarity metrics. As a
reference, the performance of the models over held-out test data (test1) is shown in Table~\ref{devel_results}, when trained both in single- and double-annotated data. We
can see that the scores for this held-out partition are better than for the final test. This could happen because we did not stratify the splits, and this test set is better aligned with the training data. In any case the difference between single and double annotation is again clear for both our classifiers.

\begin{table}[t]
  \centering
  \caption{Performance of NER systems for Mutation with pre-adjudicated annotations over held-out test data (test1).}
  \label{devel_results}
  \begin{tabular}{|lllll|}
    \hline
    \textbf{Training}&\textbf{System}    & \textbf{Precision} & \textbf{Recall}&\textbf{F-score (Std Dev)}\\ \hline
Pre-adjudicated&BioBERT & 78.1 & 63.6 & 70.1 (2.4)\\
&bilstm-crf&72.2&70.7&71.4\\
\hline
Adjudicated&BioBERT& 81.0 & 74.7 & 77.7 (2.4)\\
&bilstm-crf&85.7&72.7&78.7\\
\hline
  \end{tabular}
\end{table}

From the held-out test set, we identify all sentences that contain
at least a false positive or false negative. These sentences may
contain errors because of their similarity with inconsistently
annotated training instances, and we explore this possibility by
finding the most similar training sentences using two different
methods: Interpretable Semantic Textual Similarity (ISTS) and sentence embeddings.

We evaluate the rankings provided by the different approaches by relying on thresholds for the number of instances to be checked. There are 1,331 sentences in training data, out of which 207 (15.6\%) contain discrepancies with adjudicated ground truth. Therefore we expect a random baseline to perform with a precision close to 15.6\%. As a sanity check we perform different runs selecting random rankings from the training data, and show the results in  Table~\ref{random_ranking}. We can see that we obtain the expected scores.

\begin{table}[ht]
  \centering
  \caption{Performance of sentences identified randomly. Total sentences in training data: 1331, out of which 207 (15.6\%) contain discrepancies with adjudicated ground truth. Highest score per column in bold.}
  \label{random_ranking}
  \begin{tabular}{|lllll|}
    \hline
    \textbf{Threshold}    & \textbf{Sentences to check} & \textbf{Precision}& \textbf{Recall}&\textbf{F-score}\\
&&&&\textbf{(Std Dev)}\\
\hline
100&7.5\% & \textbf{16.0} & 8.6 & 11.2 (1.6)\\
200& 15.0\% & 14.0 & 15.5 & 14.7 (2.2)\\
500& 37.6\% & 15.9 & \textbf{41.0} & \textbf{22.8} (1.6)\\
\hline
  \end{tabular}
\end{table}

For the similarity-based methods, we rely on the errors found in the
Bilst-crf and BioBERT experiments over the held-out test set (test1), and for
each error sentence we identify the most similar sentences using ISTS
and sentence embeddings in the full training set. We measure
the ability to identify sentences where single-annotated and
double-annotated labels differ, and the results of the experiment are
given in Table~\ref{alignment}.

\begin{table}[t]
  \centering
  \caption{Performance of sentences identified by ISTS and sentence embeddings for the different NER classifiers. Total sentences in training data: 1331, out of which 207 (15.6\%) contain discrepancies with adjudicated ground truth. Highest score per column in bold.}
  \label{alignment}
  \begin{tabular}{|lllllll|}
    \hline
    \textbf{NER} & \textbf{Alignment} & \textbf{Threshold}    & \textbf{Sentences} & \textbf{Prec.}& \textbf{Rec.}&\textbf{F-sc.}\\ 
    &&&\textbf{to check}&&&\\
    \hline
Baseline&&All&1331 (100\%)&15.6&\textbf{100.0}&26.9\\
\hline
&&Top-100&7.5\%&\textbf{29.0}&14.0&18.9\\
    Bilstm-crf&ISTS&Top-200&15.0\%&24.5&23.7&24.1\\
&&Top-500&37.6\%&19.8&47.8&\textbf{28.0}\\
\hline
&&Top-100&7.5\%&19.0&9.2&12.4\\
BioBERT&ISTS&Top-200&15.0\%&20.0&19.3&19.7\\
&&Top-500&37.6\%&16.8&40.6&23.8\\
\hline
&&Top-100&7.5\%&16.0&7.7&10.4\\
BioBERT&Sentence&Top-200&15.0\%&17.5&16.9&17.2\\
&Embeddings&Top-500&37.6\%&15.4&37.2&21.8\\
\hline
  \end{tabular}
\end{table}


The results illustrate that in most cases the precision of the ranking methods improves the random baseline. The highest precision is achieved by Bilstm-crf with ISTS for the top-100 and top-200 thresholds, showing that the rankings can retrieve significantly higher amounts of useful instances than random selection. For BioBERT, we can see that ISTS performs better than sentence embeddings, which does not improve random precision.

For our second experiment on ranking evaluation, we use confidence scores, as represented
by the log probabilities of predictions of models trained on
double-annotated data. The target sentences are those in training and
development data, and for each model (Bilstm-crf and BioBERT) a ranking is
generated according to the confidence scores (cf. Figure~\ref{fig:confidence_flowchart}). The results for
different thresholds are given in Table~\ref{confidence}.

\begin{table}[ht]
  \centering
  \caption{Performance of sentences ranked by confidence for the different NER classifiers. Total sentences in training and development data: 1331, out of which 207 (15.6\%) contain discrepancies with adjudicated ground truth.}
  \label{confidence}
  \begin{tabular}{|llllll|}
    \hline
    \textbf{NER} & \textbf{Threshold}    & \textbf{Sentences} & \textbf{Precision}& \textbf{Recall}&\textbf{F-score}\\
&&\textbf{to check}&&&\\
\hline
Baseline&All&1331 (100\%)&15.6&\textbf{100.0}&26.9\\
\hline
    &Top-100&7.5\%&\textbf{34.0}&16.4&22.1\\
    Bilstm-crf&Top-200&15.0\%&32.0&30.9&31.4\\
    &Top-500&37.6\%&28.2&68.1&\textbf{39.9}\\
\hline
&Top-100&7.5\%& 32.0 & 15.5 & 20.8\\
BioBERT&Top-200&15\%& 27.5 & 26.6 & 27.0\\
&Top-500&37.6\%& 22.4 & 54.1 & 31.7\\
\hline
  \end{tabular}
\end{table}

We can see that the best results are achieved by using the confidence
scores from Bilstm-crf, and this method is able to achieve the highest
F-score for all different thresholds. This indicates that simply using
the confidence values from the Bilstm-crf prototype can help find the best instances to double-annotate in the dataset.

\subsection{Retraining models}
\label{sec:retraining}

Finally me measure the impact of building models with double-annotated
samples, and we rely on two of the ranking methods above to detect those
instances: Bilstm-crf confidence scores, and ISTS alignment over
Bilstm-crf outputs. Because of its better performance on its own, we apply BioBERT with different numbers of
double-annotated documents, and we observe whether the gap with the
full annotation has closed. The results are given in
Table~\ref{Retrain}.

\begin{table}[ht]
  \centering
  \caption{Results over test data (test2) after retraining BioBERT with confidence and ISTS-based Bilstm-crf ranking for double annotation. Best performance per column is given in bold.}
  \label{Retrain}
  \begin{tabular}{|lllll|}
    \hline
    \textbf{Ranking method}&\textbf{Threshold}    & \textbf{Precision}& \textbf{Recall}&\textbf{F-score} \\
&&&&\textbf{(Std Dev)} \\
\hline
Check none & 0 & 71.5 & 62.6 & 66.7 (2.9)\\
Check all & 1331 & 76.8 & 76.2 & 76.5 (1.1)\\
\hline
& 100 & 72.6 & 62.9 & 67.3 (2.2)\\
Random & 200 & 73.5 & 66.1 & 69.6 (2.0)\\
& 500 & 73.7 & 69.0 & 71.3 (2.0)\\
\hline
& Top-100 & 75.1 & 66.0 & 70.2 (1.9)\\
Bilstm-crf with& Top-200 & \textbf{76.0} & 68.2 & 71.9 (2.2)\\
confidence score& Top-500 & 75.4 & 68.7 & 71.9 (1.3)\\
    \hline
& Top-100 & 75.0 & 66.0 & 70.2 (2.7)\\
Bilstm-crf with ISTS& Top-200 & 75.4 & 68.4 & 71.7 (2.0)\\
& Top-500 & 74.9 & \textbf{71.0} & \textbf{72.9} (1.5)\\  
\hline
  \end{tabular}
\end{table}

We can see that the two ranking approaches (confidence-based and ISTS) perform similarly for the lower thresholds (100 and 200), and they clearly improve over random selection. For the top-500 threshold, we see that the confidence-based method does not get any improvement, while the similarity-based approach is able to gain another percentage point. This could indicate that the similarity-based method is better suited for fixing the errors of the model, even if it performs worse in the task of predicting instances that have the wrong label (Section~\ref{sec:ranking eval}). The reason for this could be that similarity-based approaches rely on the errors found on held-out data as starting point. This would make them better tuned to find sentences that have higher impact for the model, as opposed to sentences that are easier to predict as erroneous.


\section{Discussion}
\label{sec:discussion}


Our experiments show that there is a big difference in performance when the training data is annotated by one or two annotators. This effect is clear for different classifiers and test splits, and raises questions about the reliance on single-annotated data for NER in challenging domains. The performance drop takes most models from F-scores on the high 70s, to F-scores in the low 70s or 60s; this could have large effects for applications built on top of NER methods.

We explored the possibility of automatically identifying the discrepancies between pre-adjudicated and adjudicated examples by automatically ranking pre-adjudicated instances for second annotation. We proposed two scenarios: confidence-based and similarity-based. Our experiments show that the confidence-based method is able to perform with high precision (given the high bias towards negative instances), and obtain higher results than the similarity-based techniques in this task. For similarity based techniques, ISTS obtains the best result, with sentence embeddings failing to improve the random baseline.

For our last experiment we explored how the identification of errors in pre-adjudicated data translates to better NER models after re-training. We observe that in this case the biggest gains are obtained by the similarity-based technique ISTS. The reason for this could be that ISTS exploits errors over held-out data to find similar training instances, which have larger impact to build more accurate models. Another advantage of ISTS is that it provides alignment of terms and phrases in the compared sentences, which could help explain the predictions of the classifier, and make the second annotation easier for the user. By relying on ISTS, the gap in F-score for the pre-adjudicated dataset is reduced from 9.8\% to 3.6\% when adjudicating 37.6\% of the examples; when adjudicating only 100 examples (7.5\% of the training), the gap is reduced to 6.3\% for both ISTS and Bilstm-crf confidence.


\section{Conclusion}
\label{sec:conclusion}

In this article we explored the issue of the quality of manually annotated training data for NER, and the effect of using single versus double annotation per instance. Our results show a large gap in model performance when relying on the former, and we explored different ranking approaches to help choosing instances for double-annotation. We evaluated these approaches on their ability to find manual annotation differences, and also on the impact for re-training NER models. We found that confidence-based methods perform best for identifying training differences, while similarity-based techniques have the most impact for re-training NER models. Using these methods the gap between single and double-annotation can be significantly reduced without having to double-annotate the full dataset.

\section{References}

\bibliography{refs}

\end{document}